\documentclass{article}
\usepackage{spconf,amsmath,graphicx}

\usepackage{amssymb}
\usepackage{stfloats}
\usepackage{amsfonts}
\usepackage{url}
\usepackage{multirow}

\title{Exploiting Language Model for Efficient Linguistic Steganalysis}
\name{Biao Yi, Hanzhou Wu, Guorui Feng and Xinpeng Zhang
\thanks{It was supported by National Natural Science Foundation of China under Grant No. 61902235 and Shanghai ``Chen Guang'' Program under Grant No. 19CG46. Corresponding author: Hanzhou Wu (E-mail: h.wu.phd@ieee.org)}}
\address{Shanghai University, Shanghai 200444, China}

\begin{document}

\maketitle

\begin{abstract}
Recent advances in linguistic steganalysis have successively applied CNN, RNN, GNN and other efficient deep models for detecting secret information in generative texts. These methods tend to seek stronger feature extractors to achieve higher steganalysis effects. However, we have found through experiments that there actually exists significant difference between automatically generated stego texts and carrier texts in terms of the conditional probability distribution of individual words. Such kind of difference can be naturally captured by the language model used for generating stego texts. Through further experiments, we conclude that this ability can be transplanted to a text classifier by pre-training and fine-tuning to improve the detection performance. Motivated by this insight, we propose two methods for efficient linguistic steganalysis. One is to pre-train a language model based on RNN, and the other is to pre-train a sequence autoencoder. The results indicate that the two methods have different degrees of performance gain compared to the randomly initialized RNN, and the convergence speed is significantly accelerated. Moreover, our methods achieved the best performance compared to related works, while providing a solution for real-world scenario where there are more cover texts than stego texts.
\end{abstract}


\section{Introduction}
Steganography \cite{book:stego} embeds secrets in public carriers without being easily noticed by the monitor. The carrier can be generally arbitrary media. As an important carrier for people to communicate with each other in daily life, natural language is actually quite suitable for steganography, which is referred to as \emph{linguistic steganography (LS)}. The advantage is that LS can be easily concealed by the huge number of social activities.

Conventional LS mainly includes two categories: \emph{modification based} and \emph{generation based}. The former modifies a given text carrier to realize the embedding of secret information such as \cite{modification:paper1, modification:paper2, modification:paper3, modification:paper4}. Since a text is often highly coded, a significant disadvantage for modification based methods is that the maximum size of embeddable payload is small. The latter trains a language model (LM) on a corpus so that secret information can be embedded during text generation according to the LM. Compared to modification based methods, this category allows more data to be embedded. Moreover, texts generated by a well-trained LM are more semantically natural, implying that generation based methods tend to generate stego texts with higher quality. It motivates many scholars to propose generation based LS methods \cite{generation:paper1, generation:paper2, generation:paper3, generation:paper4, generation:paper5}.

\begin{figure}[!t]
\centering
\includegraphics[width=2.5in]{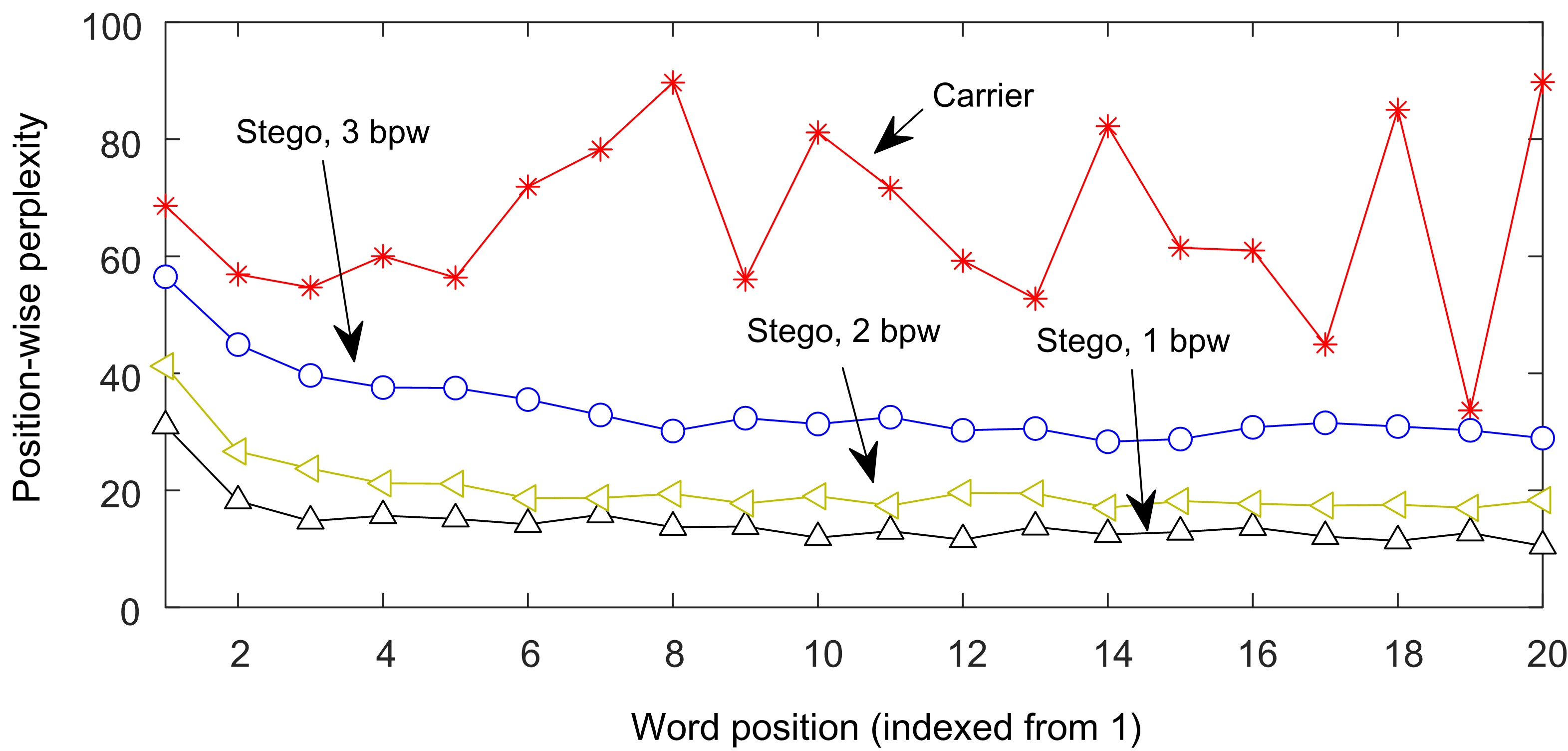}
\caption{Position-wise perplexities for carrier/stego texts.}
\end{figure}

\begin{figure}[!t]
\centering
\includegraphics[width=2.5in]{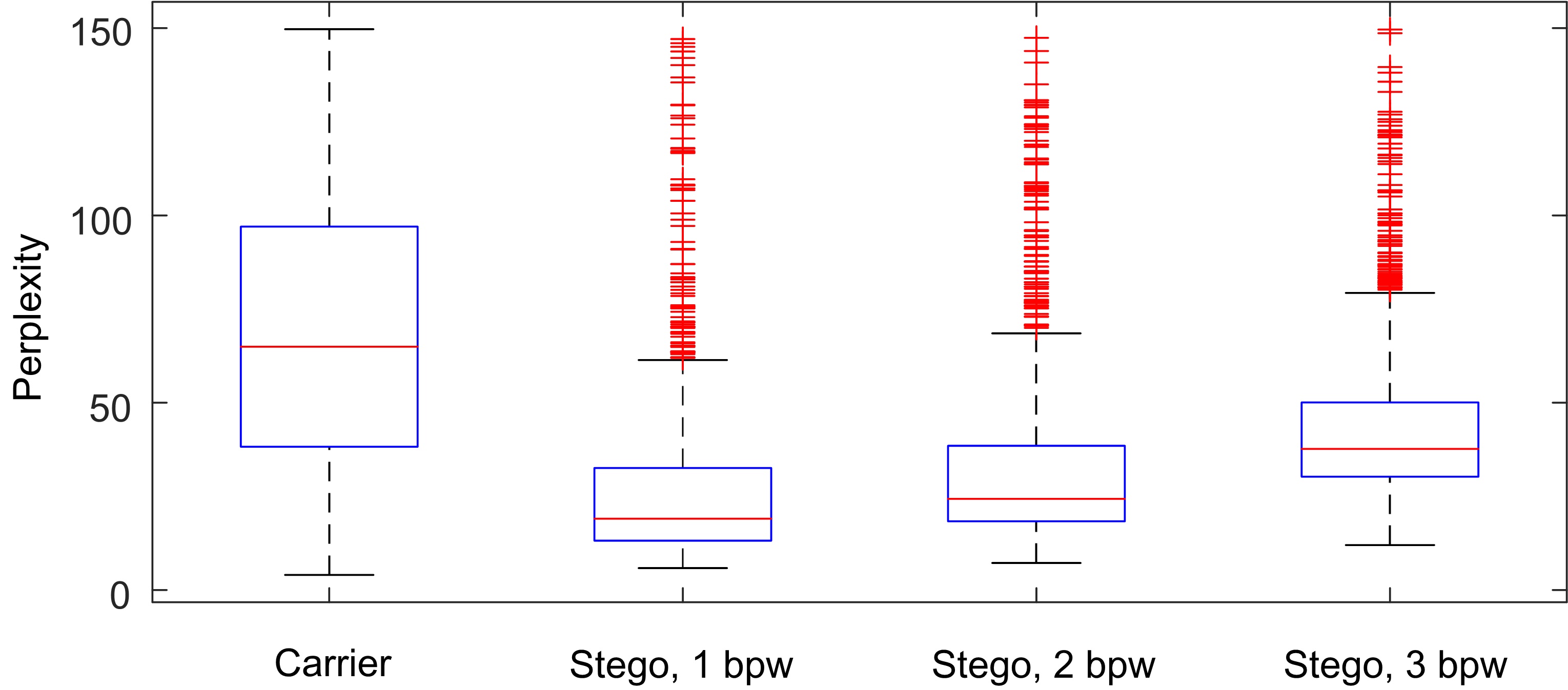}
\caption{Text-level perplexity distribution for carriers/stegos.}
\end{figure}

As the opposite of steganography, a goal of steganalysis is to detect whether there is secret data embedded in the media. We urgently need to develop a steganalysis system with efficient detection capabilities to deal with threats caused by generative LS. Early steganalysis methods \cite{EarlySteganalysis:P1, EarlySteganalysis:P2, EarlySteganalysis:P3} feed manually-crafted features into an ordinary classifier such as support vector machine for text classification. They are no longer sufficient to detect generative LS. Recently, increasing works use deep learning \cite{DL:book} for efficient linguistic steganalysis. It enables a deep neural network to automatically extract the discriminative features for classification by an end-to-end fashion. E.g., Yang \emph{et al.} \cite{Yang:SPL1} map the words in a given text to a semantic space and extract the correlation features between words using a hidden layer for final classification. Wen \emph{et al.} \cite{Wen:SPL} use convolution kernels of different sizes to extract text features to achieve steganalysis. Yang \emph{et al.} \cite{Yang:SPL2} use recurrent neural network (RNN) \cite{LSTM:Paper} to mine the distribution difference between stego texts and carrier texts for steganalysis. In recent, Wu \emph{et al.} \cite{Wu:SPL} successfully apply graph neural network (GNN) to linguistic steganalysis. 

The above methods treat linguistic steganalysis as an ordinary text classification task, focusing on improving the model architecture and finding stronger feature extractors. The consequence of this research trend is that the development of linguistic steganalysis always relies on the development of deep feature representation technologies. However, unlike LS that directly modifies given unrelated carrier texts, generative LS should pre-train a LM and then use the LM for embedding and extraction, causing the generated texts to inevitably expose the statistical characteristics to the LM, implying that uniting LM may bring us a new direction to move forward.

In this paper, we propose two methods for efficient linguistic steganalysis. One is to pre-train a language model based on RNN, and the other is to pre-train a sequence autoencoder. Experiments show that both outperform the randomly initialized RNN and the best detection performance is achieved in strict data-balanced scenario. This work has verified the effectiveness of pre-training LM for linguistic steganalysis, while firstly providing a solution for real-world scenario where there are more cover texts than stego texts.

The rest will be organized as follows. We introduce the proposed method in Section 2. Experiments and analysis are provided in Section 3. We conclude this paper in Section 4.

\section{Proposed Method}
\subsection{Motivation}
A well-trained LM used for LS naturally exposes statistical characteristics of stego texts. To explain this, we train a LM on MOVIE \cite{MOVIE:Paper} and use the method in \cite{generation:paper4} (with fixed-length coding) for generating stego texts with the trained LM. We randomly choose 1,000 stego texts generated by the trained LM for each data embedding rate and 1,000 carrier texts from the dataset. Notice that, the used LM architecture is the same as \cite{generation:paper4}. We use the well-trained LM to determine the perplexity \cite{generation:paper2} for a text $\textbf{w} =$ $(w_1, w_2, ..., w_n)$, where $n$ is the number of words. The perplexity can be expressed as:
\begin{equation}
\text{Perp}(\textbf{w}) = 2^{-\frac{1}{n}\cdot\text{log}_2\text{Pr}(w_1, w_2, ..., w_n)},
\end{equation}
where $\text{Pr}(w_1, w_2, ..., w_n) = \prod_{i=1}^{n}\text{Pr}(w_i|w_{i-1},w_{i-2},...,w_1)$ can be estimated by the LM. For each word position $i\in [1, n]$, we can also determine a perplexity as $\text{Perp}(\textbf{w}_{i}) = 2^{-\text{log}_2\text{Pr}(w_i|w_{i-1},w_{i-2},...,w_1)}$,
which is defined as \emph{position-wise perplexity}. Fig. 1 shows the (mean) position-wise perplexities. E.g., for the stego texts embedded with 1 bpw (bits per word) and a specified position, we compute the position-wise perplexity. The mean perplexity is then used as the result.

As shown in Fig. 1, there is clear difference between stego texts and carrier texts in terms of the conditional probability distribution of individual words, and the LM has the ability to characterize the difference. We show the perplexity distribution corresponding to Eq. (1) in Fig. 2. It could be inferred that we can distinguish most stego texts from normal ones by choosing a threshold (though the performance is not the best).

\begin{figure}[!t]
\centering
\includegraphics[width=2.5in]{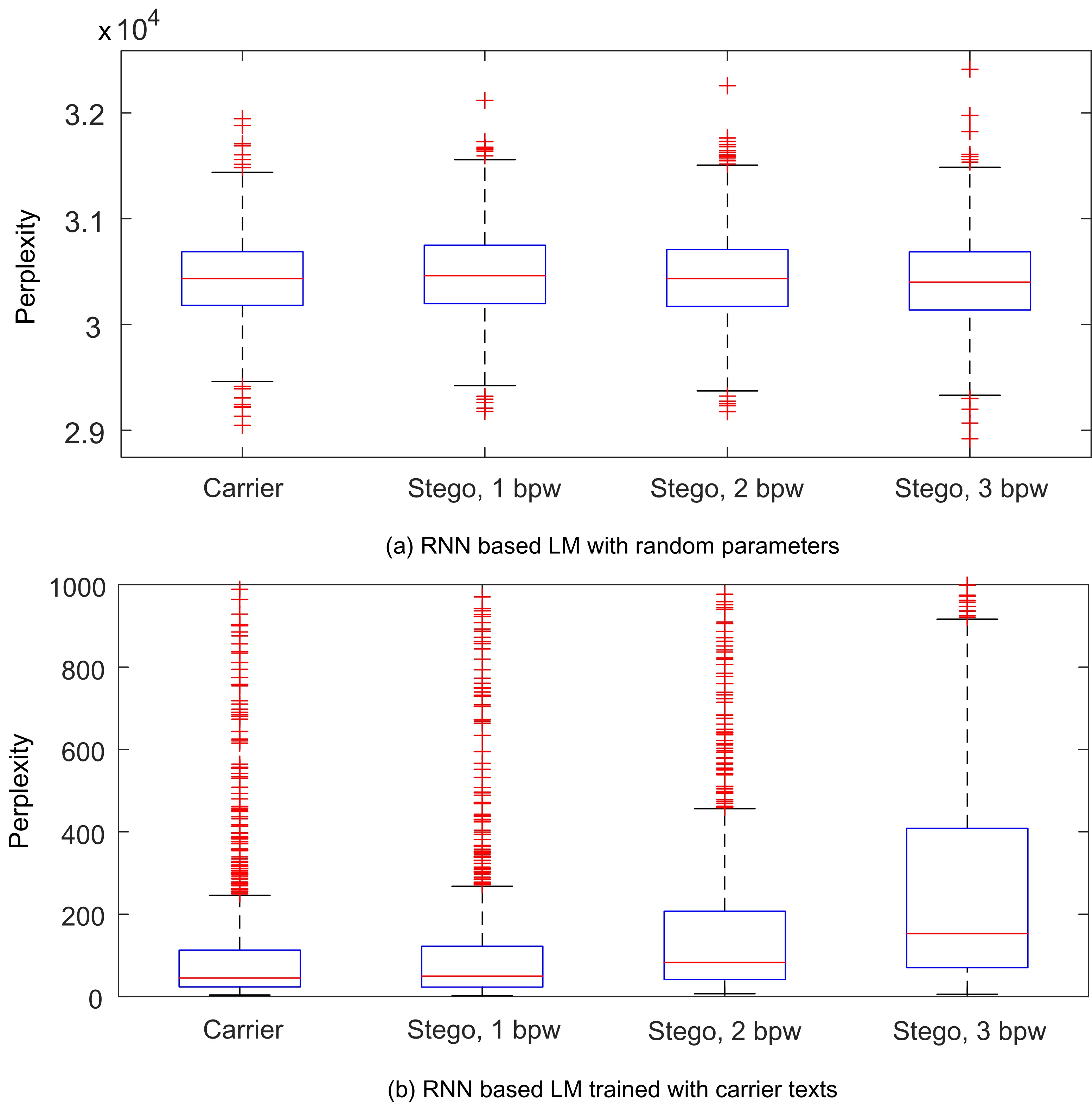}
\caption{Text-level perplexity distribution for two RNNs.}
\end{figure}

\begin{figure}[!t]
\centering
\includegraphics[width=2.5in]{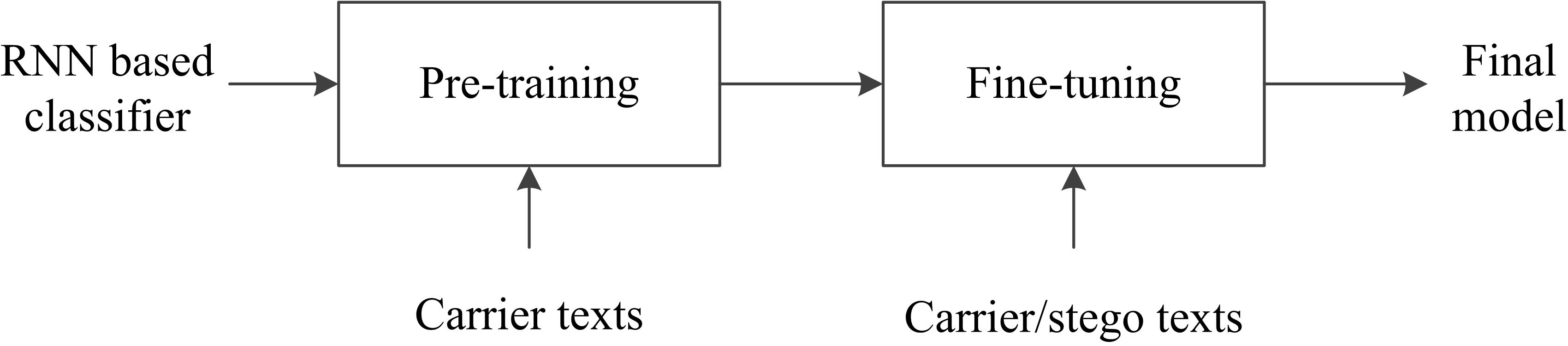}
\caption{RNN based pre-training strategy for steganalysis.}
\end{figure}

In mainstream steganalysis frameworks, the LM applied to LS is not available to the steganalyzer, who only holds some labeled carrier/stego texts for training a classifier. The practical reason is that a steganographic system is deemed secure (to a certain extent) if it manages to fool the steganalyzer even under such disadvantageous condition. It inspires us to exploit carrier texts for giving some prior knowledge to a steganalysis model by pre-training \cite{BERT:Paper, OpenAI:Paper1, Semi:Paper1}. The pre-trained parameters are used for parameter initialization for the subsequent steganalysis task to improve the detection performance.

To explain that pre-training with carrier texts indeed has the ability to capture statistical characteristics of stego texts, we use two RNNs with different parameters to represent two LMs for analyzing the perplexities. One was randomly initialized. The other was trained with 10,000 carrier texts randomly chosen from the aforementioned dataset. The LM architecture refers to Ref. \cite{generation:paper4}. Fig. 3 shows the results tested on 1,000 carriers and 1,000 stegos mentioned above. There was no intersection between carrier texts used for training and carrier texts for testing. In Fig. 3, there is no distinguishable perplexity difference between carrier texts and stego texts for the randomly initialized RNN. However, there is remarkable difference for the trained RNN, indicating that pre-training can learn prior knowledge for detecting stego texts, and thus has potential to enhance the steganalysis performance.

\subsection{Recurrent Neural Network}
We propose two pre-training methods to obtain superior steganalysis performance. One is to pre-train a traditional LM based on RNN (denoted by LM-RNN), and the other is to use a sequence autoencoder \cite{Seq2Seq:Paper} to encode a text into a vector and then produce the prediction result (denoted by AE-RNN).

RNN is the most popular neural network structure in natural language processing since its recurrent structure is suitable for processing sequences of variable lengths. The original RNN has the problem with exploding or vanishing gradients, making it difficult to learn long-term dependency. To this end, we use a variant of RNN, i.e., long and short-term memory (LSTM) \cite{LSTM:Paper}, that can effectively deal with the above problem. We refer the reader to \cite{LSTM:Paper, generation:paper2} for details about LSTM.

A common strategy of RNN is to project the hidden vector into the output space, and then use ``softmax'' \cite{DL:book} to convert it into the probabilistic space for classification or other purposes. For example, to train a LM, we project the hidden vector at each moment into the dictionary-sized output space to predict the next word. To perform steganalysis, we project the hidden state at the last moment to the output vector with a size of 2, thereby classifying a text as carrier or stego.

\begin{figure*}[!t]
\centering
\includegraphics[width=5.5in]{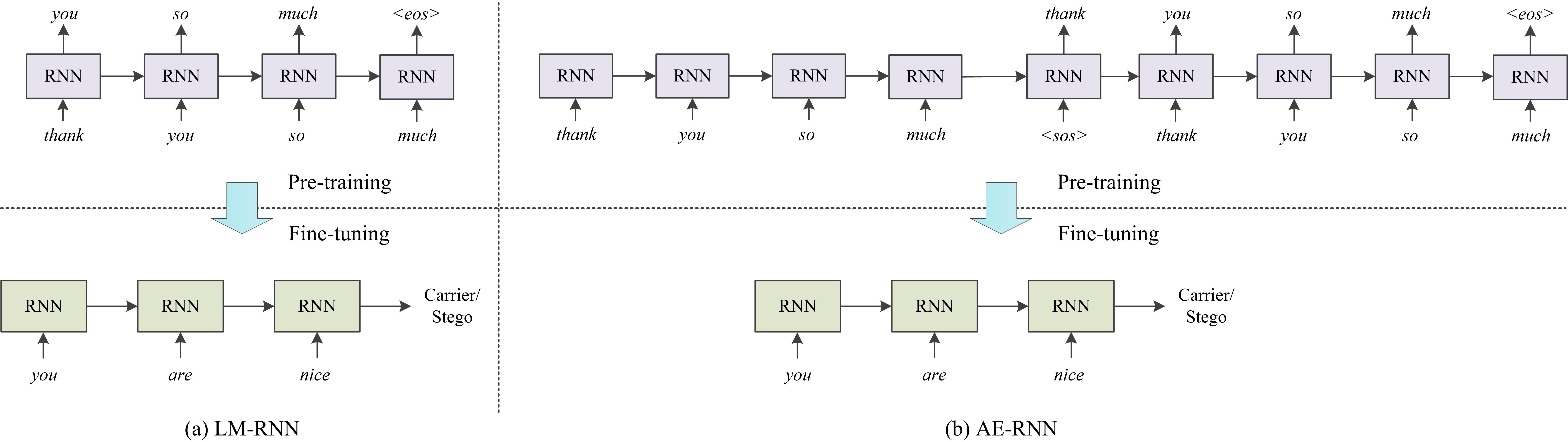}
\caption{Explanation for training (a) LM-RNN model and (b) AE-RNN model. The parameters are randomly initialized during the pre-training phase. The pre-trained parameters are then used to initialize the corresponding RNN based steganalysis model.}
\end{figure*}

\subsection{Pre-training and Linguistic Steganalysis}
We introduce two RNN based methods for steganalysis. As shown in Fig. (4, 5), both have two steps. First, carrier texts in the training set are used for pre-training. Then, the pre-trained parameters are used to initialize the model to perform regular steganalysis. Referring to Fig. 5, there are two ways for pre-training. One way is to train a traditional LM. The other is to train a sequence autoencoder, i.e., using the encoder to encode the text into a vector (that is, the hidden vector at the last moment), and then using this vector as the initial hidden vector to reconstruct the input text with the decoder. One thing to note is that the encoder and the decoder use the same LSTM network, which also means that their weights are the same.

In this study, we use the pre-trained LSTM parameters with useful prior knowledge for parameter initialization of the steganalysis model, which can improve the steganalysis performance and the convergence efficiency compared to random initialization. The specific process is briefly described as follows. We input a text into the pre-trained LSTM network and project the hidden vector at the last moment to the output vector containing two elements using a fully-connected layer. The softmax function is then used to further transform the real vector into a probability vector. During training, we minimize the cross entropy between the prediction distribution and the ground-truth distribution. During testing, a text is classified as carrier or stego based on the prediction result.

\section{Experimental Results and Analysis}
The tested LS methods include RNN-Steg \cite{generation:paper4} and Bins \cite{generation:paper1}. For RNN-Steg, two information encoding methods, i.e., fixed-length coding (FLC) and variable-length coding (VLC), were tested in the text generation stage. The details of FLC and VLC can be found in \cite{generation:paper4}. The two datasets MOVIE \cite{MOVIE:Paper} and TWITTER \cite{Twitter:Paper} were used to train LMs. The mean lengths of sentences for MOVIE and TWITTER are around 20 and 10. The numbers of sentences for them approach $1.3\times 10^6$ and $2.6\times 10^6$. Both datasets have near $5\times 10^4$ words. Each trained LM was used to produce $1\times 10^4$ stego texts with the corresponding embedding rate. These stego texts together with carrier texts (randomly chosen from the original corpus) were used for steganalysis. For each dataset, $70\%$ texts were used for training and $30\%$ for testing. In addition, $10\%$ training texts were used for validation.

For pre-training, the hyperparameters of two pre-training methods were the same. The tokenizer used in this paper was consistent with that in BERT. A word was mapped to a $128$-D vector through an embedding layer. The dropout function after the embedding layer took a retention probability of $0.5$. The number of layers of RNN was set to $2$, the hidden vector was $256$-D. The learning rate was $10^{-3}$, and the Adam \cite{Adam:Paper} optimizer was used. For pre-training, the
batch size was 128 and the number of epochs was 50. After pre-training, we directly used the pre-trained model parameters as the initial values for steganalysis training, so the hyperparameters were unchanged, but a fully connected layer was added after the last hidden vector to map it to a vector sized $2$, from which we can get a probability vector by softmax. Two common indicators \cite{Wen:SPL}: Accuracy (Acc) and F1 score (F1), were used.

\begin{table*}[!t]
\centering
\caption{Detection results using different steganalysis models under different experimental conditions.}
\scalebox{0.56}{
\begin{tabular}{c|c|c|cc|cc|cc|cc|cc|cc}
\hline
\multicolumn{3}{c|}{Steganalysis model $\mapsto$}         & \multicolumn{2}{c|}{FCN \cite{Yang:SPL1}}   & \multicolumn{2}{c|}{CNN \cite{Wen:SPL}}   & \multicolumn{2}{c|}{GNN \cite{Wu:SPL}}   & \multicolumn{2}{c|}{RNN \cite{Yang:SPL2}}   & \multicolumn{2}{c|}{LM-RNN}       & \multicolumn{2}{c}{AE-RNN}        \\ \hline
Dataset                  & LS method & bpw   & Acc & F1 & Acc & F1 & Acc & F1 & Acc & F1 & Acc    & F1     & Acc    & F1     \\ \hline
\multirow{9}{*}{TWITTER \cite{Twitter:Paper}} &        & 1.000 & 0.8142       & 0.8162      & 0.9072       & 0.9083      & 0.9125       & 0.9134      & 0.9102       & 0.9119      & 0.9082          & 0.9074          & \textbf{0.9175} & \textbf{0.9170} \\
                         & Bins \cite{generation:paper1}   & 2.000 & 0.7833       & 0.7835      & 0.8953       & 0.8977      & 0.9007       & 0.9020      & 0.8998       & 0.9027      & 0.9017          & 0.9020          & \textbf{0.9057} & \textbf{0.9068} \\
                         &        & 3.000 & 0.7647       & 0.7735      & 0.8953       & 0.8955      & 0.9013       & 0.8998      & 0.9002       & 0.8998      & 0.8988          & 0.8980          & \textbf{0.9078} & \textbf{0.9093} \\ \cline{2-15} 
                         &        & 1.000 & 0.7992       & 0.7908      & 0.8952       & 0.8966      & 0.9023       & 0.9039      & 0.9000       & 0.9015      & 0.9085          & 0.9066          & \textbf{0.9112} & \textbf{0.9120} \\
                         & FLC \cite{generation:paper4}    & 2.000 & 0.7657       & 0.7696      & 0.8937       & 0.8960      & 0.8973       & 0.8975      & 0.8820       & 0.8850      & 0.8995          & 0.9004          & \textbf{0.9035} & \textbf{0.9040} \\
                         &        & 3.000 & 0.7393       & 0.7514      & 0.8935       & 0.8937      & 0.8890       & 0.8862      & 0.8948       & 0.8988      & \textbf{0.9013} & \textbf{0.9028} & 0.8965          & 0.8996          \\ \cline{2-15} 
                         &        & 1.000 & 0.7947       & 0.7902      & 0.8943       & 0.8961      & 0.9068       & 0.9056      & 0.9043       & 0.9049      & \textbf{0.9082} & \textbf{0.9074} & 0.9043          & 0.9058          \\
                         & VLC \cite{generation:paper4}   & 2.150 & 0.7685       & 0.7708      & 0.8808       & 0.8825      & 0.8907       & 0.8872      & 0.8813       & 0.8877      & 0.8980          & 0.8969          & \textbf{0.9068} & \textbf{0.9070} \\
                         &        & 3.147 & 0.7518       & 0.7509      & 0.8842       & 0.8857      & 0.8867       & 0.8857      & 0.8877       & 0.8915      & 0.8960          & 0.8949          & \textbf{0.9027} & \textbf{0.9020} \\ \hline
\multirow{9}{*}{MOVIE \cite{MOVIE:Paper}}   &        & 1.000 & 0.8973       & 0.8935      & 0.9612       & 0.9616      & 0.9535       & 0.9541      & 0.9538       & 0.9541      & \textbf{0.9633} & \textbf{0.9635} & 0.9627          & 0.9630          \\
                         & Bins \cite{generation:paper1}  & 2.000 & 0.8575       & 0.8570      & 0.9390       & 0.9400      & 0.9428       & 0.9431      & 0.9412       & 0.9424      & \textbf{0.9515} & \textbf{0.9515} & 0.9488          & 0.9485          \\
                         &        & 3.000 & 0.8052       & 0.8032      & 0.9255       & 0.9278      & 0.9155       & 0.9148      & 0.9152       & 0.9177      & 0.9335          & 0.9339          & \textbf{0.9352} & \textbf{0.9355} \\ \cline{2-15} 
                         &        & 1.000 & 0.8848       & 0.8825      & 0.9492       & 0.9503      & 0.9540       & 0.9543      & 0.9527       & 0.9535      & 0.9592          & 0.9590          & \textbf{0.9618} & \textbf{0.9615} \\
                         & FLC \cite{generation:paper4}   & 2.000 & 0.8327       & 0.8317      & 0.9357       & 0.9368      & 0.9358       & 0.9360      & 0.9232       & 0.9268      & 0.9452          & 0.9455          & \textbf{0.9498} & \textbf{0.9504} \\
                         &        & 3.000 & 0.7815       & 0.7850      & 0.9183       & 0.9208      & 0.9210       & 0.9200      & 0.9027       & 0.9080      & 0.9302          & 0.9294          & \textbf{0.9315} & \textbf{0.9327} \\ \cline{2-15} 
                         &        & 1.000 & 0.8783       & 0.8739      & 0.9492       & 0.9498      & 0.9523       & 0.9524      & 0.9467       & 0.9479      & 0.9608          & 0.9605          & \textbf{0.9643} & \textbf{0.9645} \\
                         & VLC \cite{generation:paper4}   & 2.215 & 0.8358       & 0.8346      & 0.9307       & 0.9326      & 0.9387       & 0.9386      & 0.9330       & 0.9347      & 0.9433          & 0.9429          & \textbf{0.9475} & \textbf{0.9475} \\
                         &        & 3.260 & 0.8018       & 0.7957      & 0.9225       & 0.9242      & 0.9228       & 0.9216      & 0.9168       & 0.9201      & \textbf{0.9397} & \textbf{0.9404} & 0.9365          & 0.9365          \\ \hline
\end{tabular}}
\end{table*}

\begin{table}[!t]
\centering
\caption{Mean accuracies due to different pre-training sizes.}
\scalebox{0.65}{
\begin{tabular}{c|c|c|c}
\hline
\multicolumn{2}{c|}{Steganalysis model $\mapsto$} & \multicolumn{1}{c|}{\multirow{2}{*}{LM-RNN}} & \multicolumn{1}{c}{\multirow{2}{*}{AE-RNN}} \\
\cline{1-2}
Dataset & Number of pre-training samples  &     & \\
\hline
& $6.3\times 10^3$ & 0.9474 & 0.9487\\
MOVIE & $1.2\times 10^4$ & 0.9518 & 0.9510\\
& $1.8\times 10^4$ & 0.9535 & 0.9528\\
\hline
& $6.3\times 10^3$ & 0.9022 & 0.9062\\
TWITTER & $1.2\times 10^4$ & 0.9086 & 0.9124\\
& $1.8\times 10^4$ & 0.9120 & 0.9169\\
\hline
\end{tabular}}
\end{table}

We compare our work with GNN \cite{Wu:SPL}, fully connected network (FCN) \cite{Yang:SPL1}, convolutional neural network (CNN) \cite{Wen:SPL} and RNN \cite{Yang:SPL2}. To show the improvement brought by our models when compared to the randomly initialized RNN model, the hyperparameter settings for the RNN model used in \cite{Yang:SPL2} are consistent with ours. Table 1 shows the  results, from which we conclude that: First, the proposed LM-RNN and AE-RNN have achieved the best detection results, which can verify the feasibility and superiority of proposed work. Second, by comparing with the randomly initialized RNN \cite{Yang:SPL2}, it is inferred that the performance improvement is significant. It indicates that pre-training with carrier texts indeed improves the performance. In addition, AE-RNN is superior to LM-RNN. The reason may be that the traditional LM training focuses on predicting the next word, but AE-RNN not only does this, but also learns the entire sentence information and therefore has stronger modeling capabilities.

\begin{figure}[!t]
\centering
\includegraphics[width=2.7in]{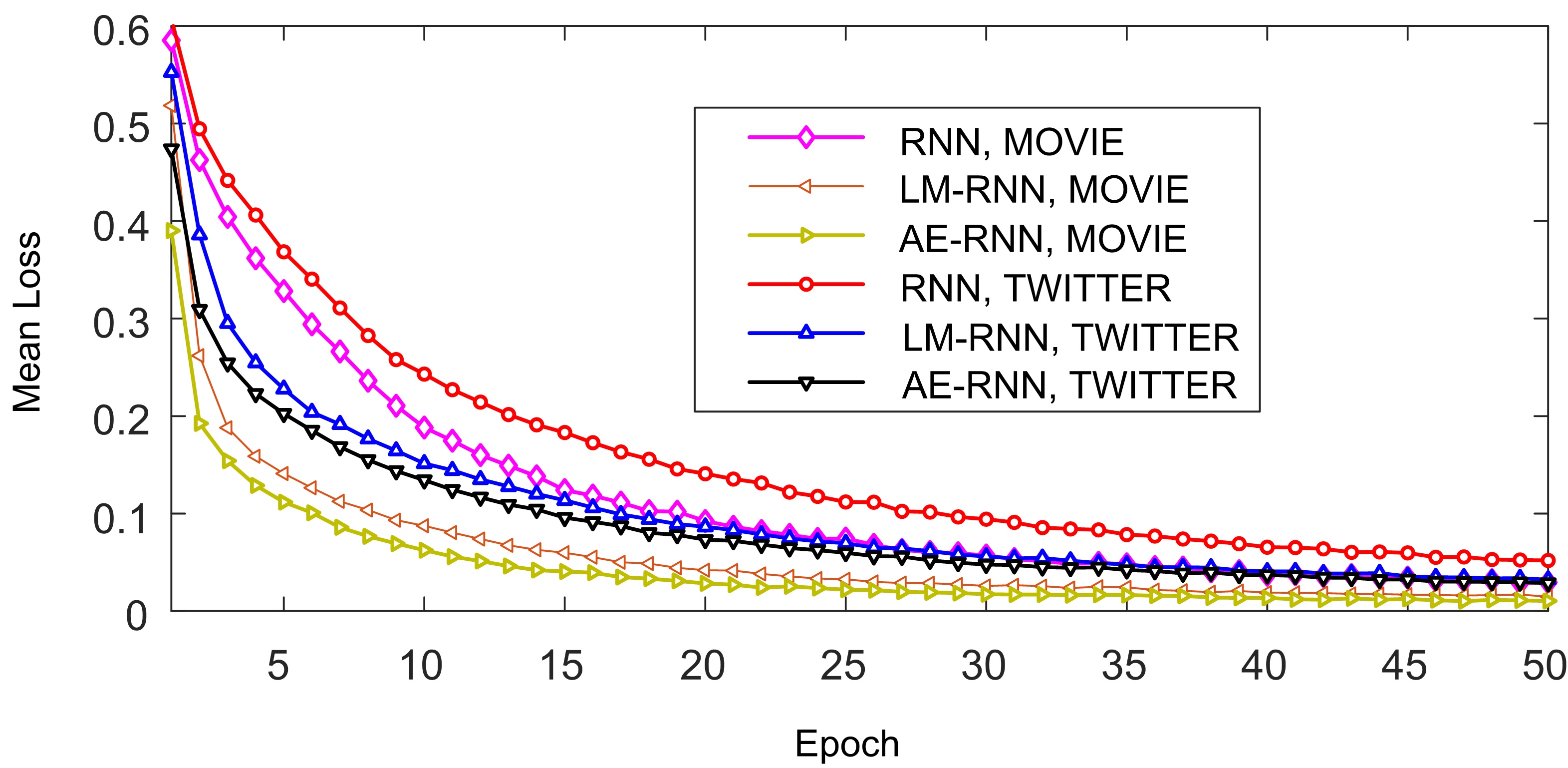}
\caption{Mean loss curves. For each point, we collect nine loss values (since there are three LS algorithms and three different embedding rates) and determine the mean value as the result.}
\end{figure}

We have also tested the performance of randomly initialized RNNs and pre-trained RNNs in terms of loss convergence efficiency. As shown in Fig. 6, pre-training leads to significant improvement. In fact, since carrier texts can be shared before performing steganalysis on the same dataset, we only need to pre-train the RNN model once to speed up various steganalysis for different steganographic methods and different embedding rates, which greatly improves the efficiency.

In practice, we may be able to collect more texts for pre-training. We use more carrier texts for pre-training to evaluate its impact on the steganalysis performance. Table 2 shows the mean accuracy values due to different sizes of the pre-training set, e.g., $0.9474$ is the mean value of nine accuracy values of LM-RNN (tested on the MOVIE dataset) shown in Table 1. There is no intersection between pre-training set and testing set. It can be inferred that the steganalysis performance can be further improved by using more carrier texts for pre-training, which inspires us to collect carrier texts as many as possible in practice so as to achieve superior steganalysis performance.

\section{Conclusion}
We found through experiments that the well-trained language model has the ability to model the distribution difference between carrier texts and steganographic texts. This ability can be passed to the classifier through pre-training and finetuning to improve the performance of subsequent steganalysis. Motivated by this important insight, we propose two methods for enhancing the performance of RNN-based steganalysis models. The experimental results have shown that the two methods have achieved the best performance compared to related works and can greatly improve the training efficiency.

\end{document}